\documentclass[letterpaper]{article} 
\usepackage{aaai23}  
\usepackage{times}  
\usepackage{helvet}  
\usepackage{courier}  
\usepackage[hyphens]{url}  
\usepackage{graphicx} 
\usepackage{natbib}  
\usepackage{caption} 
\usepackage{algorithm}
\usepackage{algorithmic}

\usepackage{newfloat}
\usepackage{listings}
\DeclareCaptionStyle{ruled}{labelfont=normalfont,labelsep=colon,strut=off} 
\floatstyle{ruled}
\newfloat{listing}{tb}{lst}{}
\floatname{listing}{Listing}

\copyrighttext{Presented at the AI-HRI Symposium at AAAI Fall Symposium Series (FSS) 2022}

\title{A Symbolic Representation of Human Posture for Interpretable Learning and Reasoning}
\author {
    Richard G. Freedman,\textsuperscript{\rm 1}
    Joseph B. Mueller,\textsuperscript{\rm 1} 
    Jack Ladwig,\textsuperscript{\rm 1}
    Stephen Johnston,\textsuperscript{\rm 1}
    David McDonald,\textsuperscript{\rm 1}
    Helen Wauck,\textsuperscript{\rm 1}
    Ruta Wheelock,\textsuperscript{\rm 1}
    Hayley Borck\textsuperscript{\rm 1}
}
\affiliations {
    \textsuperscript{\rm 1} SIFT\\
    rfreedman@sift.net, jmueller@sift.net, jladwig@sift.net, sjohnston@sift.net, dmcdonald@sift.net, hwauck@sift.net, rwheelock@sift.net, hborck@sift.net
}

\usepackage{bibentry}

\usepackage{amsmath}
\usepackage{amssymb}
\usepackage{multirow}
\usepackage{subcaption}
\usepackage{bm}

\begin{document}

\maketitle

\begin{abstract}
Robots that interact with humans in a physical space or application need to think about the person's posture, which typically comes from visual sensors like cameras and infra-red.  Artificial intelligence and machine learning algorithms use information from these sensors either directly or after some level of symbolic abstraction, and the latter usually partitions the range of observed values to discretize the continuous signal data.  Although these representations have been effective in a variety of algorithms with respect to accuracy and task completion, the underlying models are rarely interpretable, which also makes their outputs more difficult to explain to people who request them.  Instead of focusing on the possible sensor values that are familiar to a machine, we introduce a qualitative spatial reasoning approach that describes the human posture in terms that are more familiar to people.  This paper explores the derivation of our symbolic representation at two levels of detail and its preliminary use as features for interpretable 
activity recognition.
\end{abstract}

\section{Introduction}
The number of situations in which robots interact with people's physical presence is increasing, from working in the same factory space \cite{DBLP:conf/case/LasotaRS14} to dance partners \cite{irr_danceRobot}.  Working together in these conditions involves coordination that requires reasoning about how people pose and move \cite{DBLP:conf/humanoids/MaedaELA0N14,DBLP:conf/icra/MainpriceHB15,Mortl01112012_movingTable,7140067_pathPlanHumanPrediction}.  Even without direct physical proximity, robots often need to think about a person's physical posture in applications such as sports coaching \cite{sportsCoachRobot} and elder care, which spans tasks from exercise engagement \cite{mataric_elderCareRobot} to fall detection \cite{7333644_DBMMactivityRecog,fallDetectionRobot_6705510,BEDDIAR2022103407_fallDetect}.  In motion planning, robots need to further think about their own embodiment and posture to accomplish tasks; learning from human demonstrations via observation \cite{DBLP:conf/iros/InamuraNS02_mirrorNeuron,DBLP:conf/icra/InamuraTN02_mimesis} has to map human motion into the robot's analogous configurations.  It is now often the case that intelligent human-robot interaction (HRI) must consider a person's body as much as their mind.

There is plenty of research on how to use information about a person's posture for artificial intelligence (AI) and machine learning (ML).  The majority of this work involves integrating real-world perception of people with the system's internal representation of the world; that is, interpreting data from camera-like sensors.  While early efforts analyzed properties of the pixelated image, such as identifying geometrical elements \cite{NIPS2006_a209ca7b,DBLP:conf/cvpr/BissaccoYS07} and flow fields \cite{Schunck1988}, the widespread commercialization of red-green-blue-depth (RGB-D) sensors like the Kinect camera added additional dimensions (primarily depth) to the representation possibilities \cite{rgbd-survey}.  Specifically, pose estimation \cite{pose-estimation-survey} in three-dimensional space became as plausible as raw signal processing---robots with these sensors can use pixels and depth values, mesh nets that describe the shape of identified humans, or coordinates that construct stick figures of identified humans.  The latter two 
introduce layers of abstraction to the signal data, assigning 
semantic information that summarizes 
the content in the stream of sensor readings.

With many off-the-shelf algorithms available for AI and ML, the research directions when robots need to think about a person's physical posture are usually more focused on data representation to use these existing algorithms rather than creating original algorithms.  We will review some of the representations that researchers have developed in Section~\ref{sec:background}, but a common trend among them is a focus on discretizing and/or compressing the large space of possible sensor values regardless of the level of abstraction.  These are both important because the majority of such algorithms assume that input data is discrete and there is generally a need for ``similarity'' between related input instances (affecting compression choices to enforce this property).  With a focus on successful task completion or high accuracy scores, the definition of ``similarity'' traditionally defaults to some form of clustering or numeric interval partitioning; both are systematic to facilitate optimizing performance when tuning (hyper)parameters for both the algorithm and compression.

The rise of explainable AI-HRI \cite{aihri-explainableTheme} brings systematic methods such as these into question because the robot's justification for some outcome will rely on all aspects of the representation choice.  Most people do not think in pixels, depth values, 
coordinates, or other quantitative values when describing the human posture.  Presenting collections of that information based on whatever parameter gave the best results risks further obfuscation of an unnatural perspective.  Even if the algorithm has an effective means of explanation, the references to the input data (human posture, in this case) might not make sense; the robot could sound like it is speaking a foreign language to the average person in the same way a domain-expert does when using niche jargon.  This permeates to interpretability of the models these algorithms use or develop \cite{rudin_nmi19} because the structure of knowledge builds on the input representation.

We propose a novel representation of human posture data that qualitatively describes a 
stick figure 
with respect to words people commonly use (for example, `bent' and `raised').  The descriptions are symbolic, serving as logical fluents or features depending on the AI or ML algorithm. 
Section~\ref{sec:approach} introduces our representation as well as the qualitative spatial reasoning derivations. 
We then provide example applications of our new representation for activity recognition via both expert systems and case-based reasoning.  The empirical results in Section~\ref{sec:evaluation} provide evidence that the models are interpretable, and we discuss how this enables the algorithms to also be explainable. 
As this work is still in preliminary stages, Section~\ref{sec:conclusions} includes our future directions alongside a discussion about how the AI-HRI community can apply similar efforts to 
their own 
data representations.

\section{Related Works\label{sec:background}}
Prior to deep neural networks, which generally receive the raw image as input, many AI and ML algorithms required an alternative representation of the image as input that reduced the size of the input space.  For sensor data containing a human's posture, it was relatively common to perform some form of preprocessing that made the data compatible with the algorithm of choice.  Some would annotate the posture as flow fields \cite{poseFromFlowFields,WangM09} or sets of coordinates forming a stick figure \cite{poseBasedActivityRecog,kinectPoseDetection,er_isr21}.  Others compressed the entirety of the sensor data based on select features such as spatio-temporal features \cite{6094489:4DhumanActivityRecog}; the present-day analogue of this approach is to reduce the dimensionality of the data with a variational autoencoder \cite{nvae_deepCompressImage}.  As a mix of the two, some works compressed the space of annotated postures by partitioning the continuous space into discrete bins; these qualitative abstractions either captured the pose within a single frame \cite{fjz_icaps2014,fjz_fss2015,cf_aaai2018} or described relations between parts of the body and environment \cite{dachc_aaai17,dhc_aij19}.  The representations we propose fall into this hybrid category, but we focus on abstractions that are defined with respect to human-interpretable semantics \cite{explicitSemanticAnalysis_wikipedia,krs_nips2014,cltbrs_nips19} rather than systematic partitions of the continuous space.

\section{Qualitative Posture Representation\label{sec:approach}}
Our methods assume that there is a stick figure representation of the person, which is a collection of labeled coordinates in $\mathbb{R}^{3}$ representing joints in the human body (e.g.~shoulder, wrist, hip).  Although it is common for robots to generate a stick figure with RGB-D data that requires a sensor similar to the Kinect, recent advances in deep learning have led to pose detection algorithms that only need RGB data \cite{tensorflowpose,openpose,blazepose}.  We used BlazePose \cite{blazepose} within the MediaPipe \cite{mediapipe} suite to perform three-dimensional pose estimation with video on a smartphone camera, but this is not a requirement to use or replicate our work.  Different pose detection algorithms might output different sets of joints or label them differently, which would affect the available symbols one can compute.  We group the available joints into specific pairs that form links in the human body, such as upper/lower appendages and the waist.  In our derivations, we use vector notation to represent links.

\subsection{Low Level: Fluent Descriptions\label{sec:approach.descriptors}}
Given the joints available from BlazePose, we defined the logic fluents below to describe human posture.  Each fluent includes a set of complements specifying a joint's or link's possible states, where a transition between states depends on how we interpret the numerical information.  Additionally, each fluent has both an instantaneous (still image) version and a locally temporal (change over the past few frames of motion) version.

\begin{figure}
  \centering
  \includegraphics[scale=0.31]{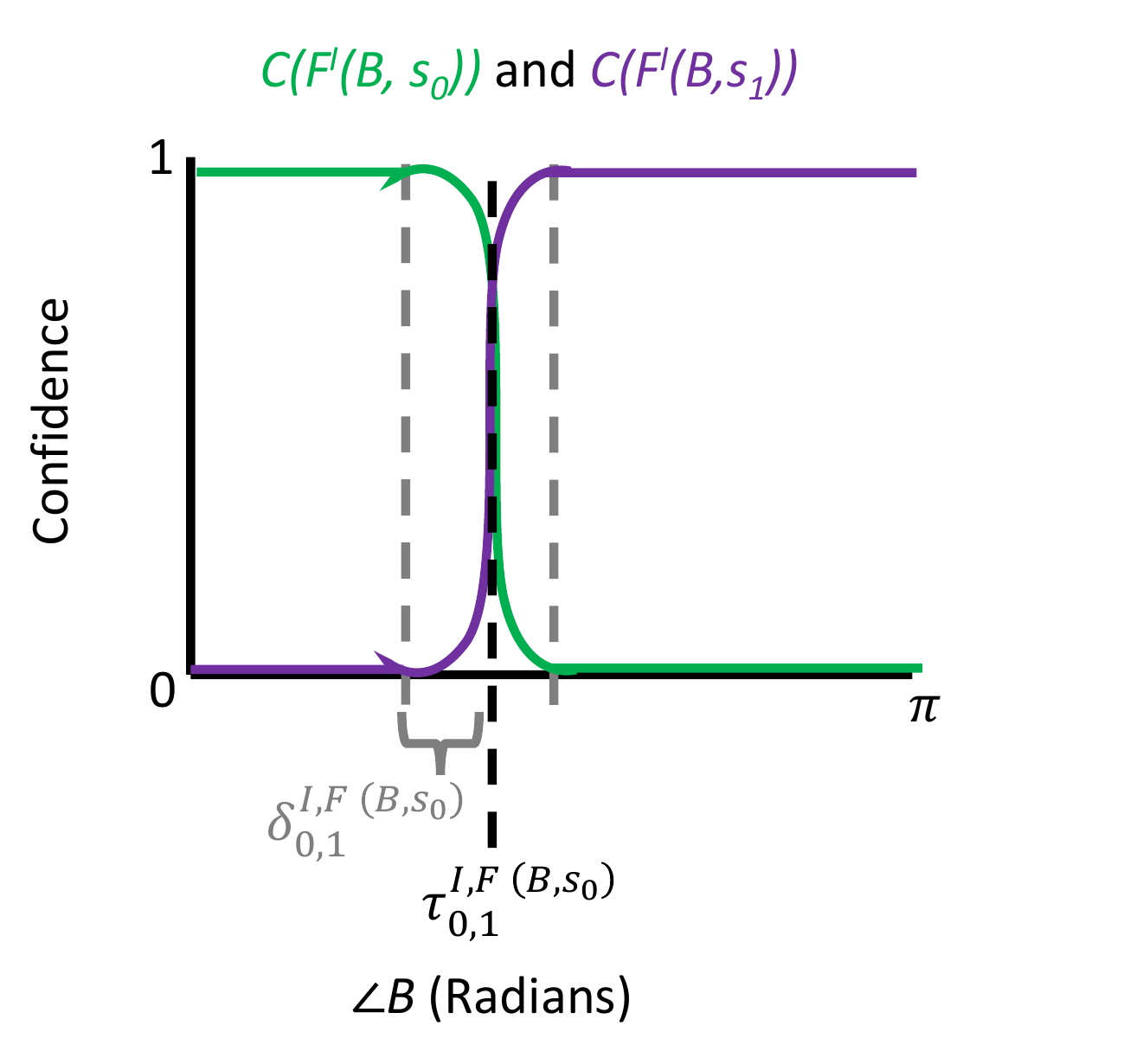}
  \includegraphics[scale=0.31]{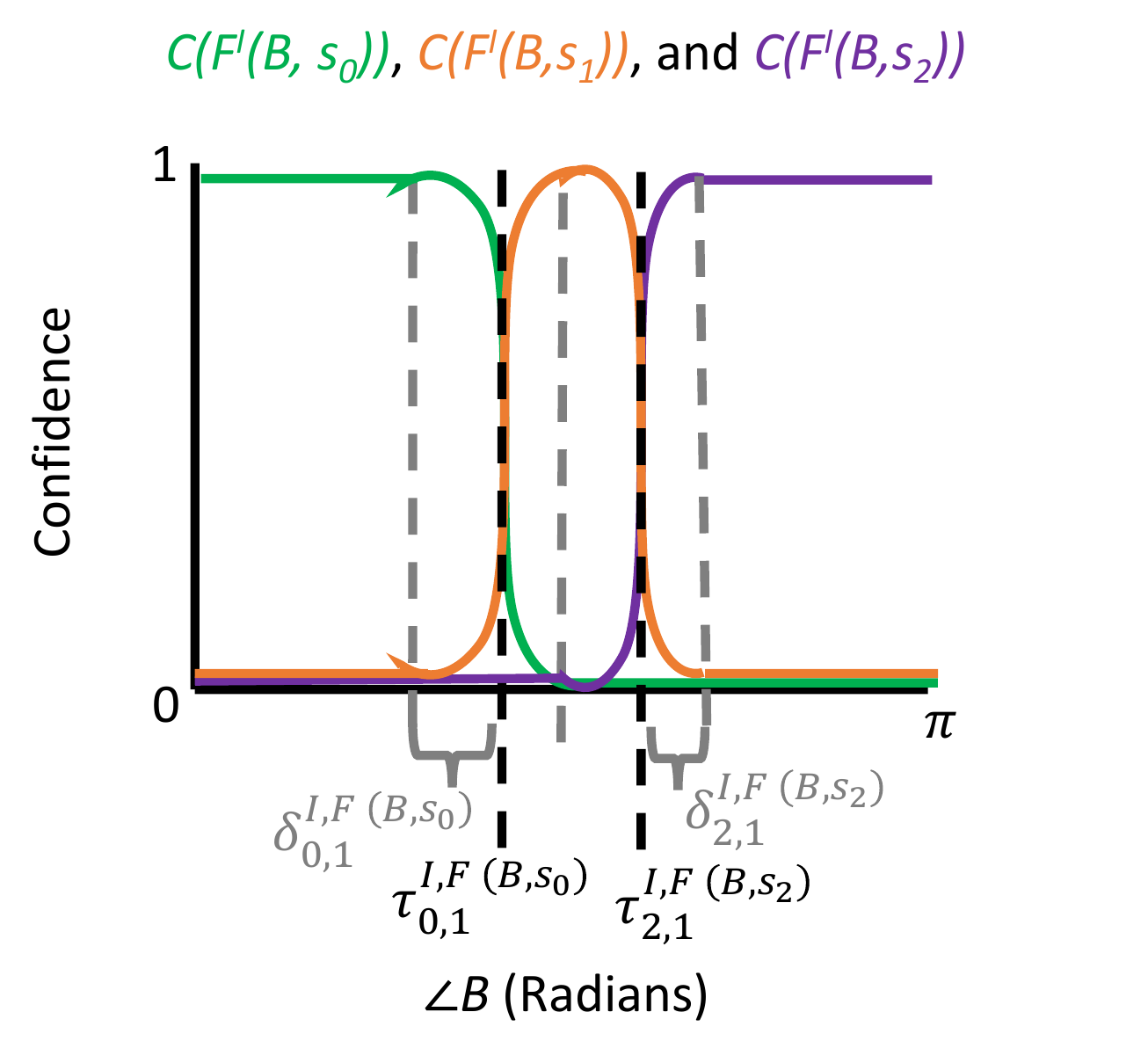}
  \caption{Confidence of fluent with respect to the angle. \label{fig:instantaneousConfidence}}
\end{figure}

The instantaneous version's derivations apply the Law of Cosines to determine some angle, forming a triangle between two connected links (sometimes projected onto a plane).  For example, a bent or straight arm measures the angle of the respective elbow.  If we have links $\overrightarrow{AB}$ and $\overrightarrow{BC}$ that share joint $B$, then with respect to triangle $\triangle ABC$, the Law of Cosines states that 
\begin{equation}\label{eq:lawOfCos}
  \angle B = \cos^{-1} \left( \frac{ \left|\overrightarrow{A,B}\right|^2 + \left|\overrightarrow{B,C}\right|^2 - \left|\overrightarrow{C,A}\right|^2 } {2 \left|\overrightarrow{A,B}\right| \left|\overrightarrow{B,C}\right| } \right) .
\end{equation}
In general, the angle determines the instantaneous fluent $F^{I}\left(\cdot\right)$'s value based on how it compares to the angle threshold(s) $\tau^{I,F\left(\cdot\right)}_{i,i+1}$ or $\tau^{I,F\left(\cdot\right)}_{i+1,i}$ between states $s_{0}, s_{1}, \ldots, s_{n}$.  However, the thresholds are approximate, 
and people might express uncertainty or disagreement of the state when the angle is close to a threshold (within $\delta^{I,F\left(\cdot\right)}_{i,i+1}$ or $\delta^{I,F\left(\cdot\right)}_{i+1,i}$).  Table~\ref{tab:thresholdVals} lists the parameter values we used in this work.  We capture this uncertainty with a confidence value $C$ that is usually $0$ or $1$, but interpolates with a logistic curve around these threshold(s) (see Figure~\ref{fig:instantaneousConfidence}):
\begin{align}\label{eq:confInstantaneous_lowerbound}
  \begin{split}
  C & \left(F^{I}\left(B,s_{i}\right)\right) = \\
    & \left\lbrace\begin{array}{rcl}
  1 & \multicolumn{2}{r}{\textnormal{if } \angle B < \tau^{I,F\left(B,s_{i}\right)}_{i,i+1} - \delta^{I,F\left(B,s_{i}\right)}_{i,i+1}} \\
  0 & \multicolumn{2}{r}{\textnormal{if } \angle B > \tau^{I,F\left(B,s_{i}\right)}_{i,i+1} + \delta^{I,F\left(B,s_{i}\right)}_{i,i+1}} \\
  \multicolumn{2}{r}{\left(1+e^{\frac{\log\left(\frac{1-\epsilon}{\epsilon}\right)}{\delta^{I,F\left(B,s_{i}\right)}_{i,i+1}}\left(\angle B-\tau^{I,F\left(B,s_{i}\right)}_{i,i+1}\right)}\right)^{-1}} & \textnormal{otherwise} \\
  \end{array}\right.
  \end{split}
\end{align}
\begin{align}\label{eq:confInstantaneous_upperbound}
  \begin{split}
  C & \left(F^{I}\left(B,s_{i}\right)\right) = \\
    & \left\lbrace\begin{array}{rcl}
  0 & \multicolumn{2}{r}{\textnormal{if } \angle B < \tau^{I,F\left(B,s_{i}\right)}_{i+1,i} - \delta^{I,F\left(B,s_{i}\right)}_{i+1,i}} \\
  1 & \multicolumn{2}{r}{\textnormal{if } \angle B > \tau^{I,F\left(B,s_{i}\right)}_{i+1,i} + \delta^{I,F\left(B,s_{i}\right)}_{i+1,i}} \\
  \multicolumn{2}{r}{\left(1+e^{\frac{\log\left(\frac{1-\epsilon}{\epsilon}\right)}{-\delta^{I,F\left(B,s_{i}\right)}_{i+1,i}}\left(\angle B-\tau^{I,F\left(B,s_{i}\right)}_{i+1,i}\right)}\right)^{-1}} & \textnormal{otherwise} \\
  \end{array}\right.
  \end{split}
\end{align}
where $\epsilon$ is a very small value to avoid dividing by $0$.  
When there are only two possible states for an instantaneous fluent, the sum of the confidence values is clearly $1$.  To preserve this property with three possible states, we compute the middle state's confidence relative to the other states' confidence:
\begin{align}\label{eq:confInstantaneous_sumTo1}
  \begin{split}
    C & \left(F^{I}\left(B,s_{1}\right)\right) = \\
    & \left(1 - C\left(F^{I}\left(B,s_{0}\right)\right)\right) \left(1 - C\left(F^{I}\left(B,s_{2}\right)\right)\right) \\
  \end{split}
\end{align}
Equations~\ref{eq:confInstantaneous_lowerbound} and~\ref{eq:confInstantaneous_upperbound} assume that the states are sequential semantically.  If a fluent describes 
a property that does not satisfy this assumption, such as whether the elbow is at a right angle (if considered binary rather than ``not yet,'' ``right angle,'' or ``too much''), then these equations will 
need modification for additional state transitions.

The locally temporal versions assess their instantaneous counterpart's measured angle over a sliding window of consecutive video frames to detect the rate of change, which is compared to thresholds $\tau^{LT,F\left(\cdot\right)}_{i,i+1} \pm \delta^{LT,F\left(\cdot\right)}_{i,i+1}$.  The locally temporal fluent $F^{LT}\left(\cdot\right)$ has 
states that indicate motion towards the two most extreme $F^{I}\left(\cdot\right)$ states $\rightarrow s_{0}$ and $\rightarrow s_{n}$; there is also a state for no motion at all $s_{\emptyset}$.  The locally temporal variant of a bent or straight arm, bending or straightening, considers the angle of the elbow in the past $r$ recorded frames.  As an approximation of the derivative of the angle over time, we simply subtract the angle's measurement between consecutive frames.  In order to focus on the more recent changes for greater relevance, the confidence that the approximate derivative indicates motion is raised to a power inversely proportional to its age (bringing older angle change comparisons closer to the multiplicative identity of $1$):
\begin{align}\label{eq:confLocalTemp_dir}
  \begin{split}
    & C\left(F^{LT}\left(B,\rightarrow s_{i}\right) \left| \textnormal{frame } f, \textnormal{window size } r \right.\right) = \\
    & \prod_{t = f - r + 1}^{f}
    C\left(F^{I}\left(\angle B_{t} - \angle B_{t-1}, \rightarrow s_{i}\right)\right)^{\frac{1}{f - t + 1}} \\
  \end{split}
\end{align}
where the instantaneous fluent version of $\left(\angle B_{t} - \angle B_{t-1}\right)$ has two states, $s_{0}^{\rightarrow} = \rightarrow s_{i}$ and $s_{1}^{\rightarrow} = \not\rightarrow s_{i}$, and uses the locally temporal threshold parameters above.  We kept the sliding window at a constant $r = 5$ for all locally temporal fluents, and the thresholds are constant $\tau^{LT,F\left(B,\rightarrow s_{i}\right)}_{0,1} = \pm \pi/72$ (depending on whether the direction is positive or negative) and $\delta^{LT,F\left(B,\rightarrow s_{i}\right)}_{i,i+1} = \pi/72$ for all locally temporal fluents that use angle measurements. The confidence that there is no motion towards any state depends on the confidence values for motion towards each extreme state:
\begin{align}\label{eq:confLocalTemp_none}
  \begin{split}
    & C \left(F^{LT}\left(B,s_{\emptyset}\right) \left| \cdot \right.\right) = \\ 
   & \left(1 - C\left(F^{LT}\left(B,\rightarrow s_{0}\right) \left| \cdot \right.\right)\right) \left(1 - C\left(F^{LT}\left(B,\rightarrow s_{n}\right) \left| \cdot \right.\right)\right) \\
  \end{split}
\end{align}

\begin{table*}
  \centering
  \caption{Threshold Values for Angle-Derived Fluent State Transitions ($\cdot$ Denotes Left or Right Side)\label{tab:thresholdVals}}
  \begin{tabular}{|c|c|c|}
    \hline
    arm bent & $\tau^{I,F\left(\cdot\textnormal{ elbow},\textnormal{bent}\right)}_{0,1} = 5\pi/8$ & $\delta^{I,F\left(\cdot\textnormal{ elbow},\textnormal{bent}\right)}_{0,1} = \pi/8$ \\
    \hline
    leg bent & $\tau^{I,F\left(\cdot\textnormal{ knee},\textnormal{bent}\right)}_{0,1} = 19\pi/24$ & $\delta^{I,F\left(\cdot\textnormal{ knee},\textnormal{bent}\right)}_{0,1} = \pi/24$ \\
    \hline
    \hline
    arm in front & $\tau^{I,F\left(\cdot\textnormal{ upper arm (sagittal)},\textnormal{in front}\right)}_{0,1} = 7\pi/16$ & $\delta^{I,F\left(\cdot\textnormal{ upper arm (sagittal)},\textnormal{in front}\right)}_{0,1} = \pi/16$ \\
    \hline
    arm behind & $\tau^{I,F\left(\cdot\textnormal{ upper arm (sagittal)},\textnormal{behind}\right)}_{2,1} = 9\pi/16$ & $\delta^{I,F\left(\cdot\textnormal{ upper arm (sagittal)},\textnormal{behind}\right)}_{2,1} = \pi/16$ \\
    \hline
    leg in front & $\tau^{I,F\left(\cdot\textnormal{ upper leg (sagittal)},\textnormal{in front}\right)}_{0,1} = 7\pi/16$ & $\delta^{I,F\left(\cdot\textnormal{ upper leg (sagittal)},\textnormal{in front}\right)}_{0,1} = \pi/16$ \\
    \hline
    leg behind & $\tau^{I,F\left(\cdot\textnormal{ upper leg (sagittal)},\textnormal{behind}\right)}_{2,1} = 9\pi/16$ & $\delta^{I,F\left(\cdot\textnormal{ upper leg (sagittal)},\textnormal{behind}\right)}_{2,1} = \pi/16$ \\
    \hline
    \hline
    arm raised & $\tau^{I,F\left(\cdot\textnormal{ armpit},\textnormal{raised}\right)}_{2,1} = 9\pi/16$ & $\delta^{I,F\left(\cdot\textnormal{ armpit},\textnormal{raised}\right)}_{2,1} = \pi/16$ \\
    \hline
    arm lowered & $\tau^{I,F\left(\cdot\textnormal{ armpit},\textnormal{lowered}\right)}_{0,1} = 3\pi/8$ & $\delta^{I,F\left(\cdot\textnormal{ armpit},\textnormal{lowered}\right)}_{0,1} = \pi/8$ \\
    \hline
    leg raised & $\tau^{I,F\left(\cdot\textnormal{ hip},\textnormal{raised}\right)}_{0,1} = 13\pi/16$ & $\delta^{I,F\left(\cdot\textnormal{ hip},\textnormal{raised}\right)}_{0,1} = \pi/16$ \\
    \hline
    \hline
    arm outward & $\tau^{I,F\left(\cdot\textnormal{ armpit (frontal)},\textnormal{outward}\right)}_{0,1} = 7\pi/16$ & $\delta^{I,F\left(\cdot\textnormal{ armpit (frontal)},\textnormal{outward}\right)}_{0,1} = \pi/16$ \\
    \hline
    arm inward & $\tau^{I,F\left(\cdot\textnormal{ armpit (frontal)},\textnormal{inward}\right)}_{2,1} = 9\pi/16$ & $\delta^{I,F\left(\cdot\textnormal{ armpit (frontal)},\textnormal{inward}\right)}_{2,1} = \pi/16$ \\
    \hline
    leg outward & $\tau^{I,F\left(\cdot\textnormal{ hip (frontal)},\textnormal{outward}\right)}_{0,1} = 7\pi/16$ & $\delta^{I,F\left(\cdot\textnormal{ hip (frontal)},\textnormal{outward}\right)}_{0,1} = \pi/16$ \\
    \hline
    leg inward & $\tau^{I,F\left(\cdot\textnormal{ hip (frontal)},\textnormal{inward}\right)}_{2,1} = 9\pi/16$ & $\delta^{I,F\left(\cdot\textnormal{ hip (frontal)},\textnormal{inward}\right)}_{2,1} = \pi/16$ \\
    \hline
    \hline
    head tilted forward & $\tau^{I,F\left(\textnormal{neck (sagittal)},\textnormal{tilted forward}\right)}_{0,1} = 7\pi/16$ & $\delta^{I,F\left(\textnormal{neck (sagittal)},\textnormal{tilted forward}\right)}_{0,1} = \pi/16$ \\
    \hline
    head tilted left & $\tau^{I,F\left(\textnormal{neck (frontal)},\textnormal{tilted left}\right)}_{0,1} = 7\pi/16$ & $\delta^{I,F\left(\textnormal{neck (frontal)},\textnormal{tilted left}\right)}_{0,1} = \pi/16$ \\
    \hline
    torso tilted forward & $\tau^{I,F\left(\textnormal{mid-hip (sagittal)},\textnormal{tilted forward}\right)}_{0,1} = 7\pi/16$ & $\delta^{I,F\left(\textnormal{mid-hip (sagittal)},\textnormal{tilted forward}\right)}_{0,1} = \pi/16$ \\
    \hline
    torso tilted left & $\tau^{I,F\left(\textnormal{mid-hip (frontal)},\textnormal{tilted left}\right)}_{0,1} = 7\pi/16$ & $\delta^{I,F\left(\textnormal{mid-hip (frontal)},\textnormal{tilted left}\right)}_{0,1} = \pi/16$ \\
    \hline
    \hline
    head twisted left & $\tau^{I,F\left(\textnormal{neck (transverse)},\textnormal{twisted left}\right)}_{0,1} = 7\pi/16$ & $\delta^{I,F\left(\textnormal{neck (transverse)},\textnormal{twisted left}\right)}_{0,1} = \pi/16$ \\
    \hline
    torso twisted left & $\tau^{I,F\left(\textnormal{mid-hip (transverse)},\textnormal{twisted left}\right)}_{0,1} = 7\pi/16$ & $\delta^{I,F\left(\textnormal{mid-hip (transverse)},\textnormal{twisted left}\right)}_{0,1} = \pi/16$ \\
    \hline
  \end{tabular}
\end{table*}

\begin{figure*}
  \centering
  \begin{subfigure}[b]{0.25\textwidth}
    \includegraphics[width=\textwidth]{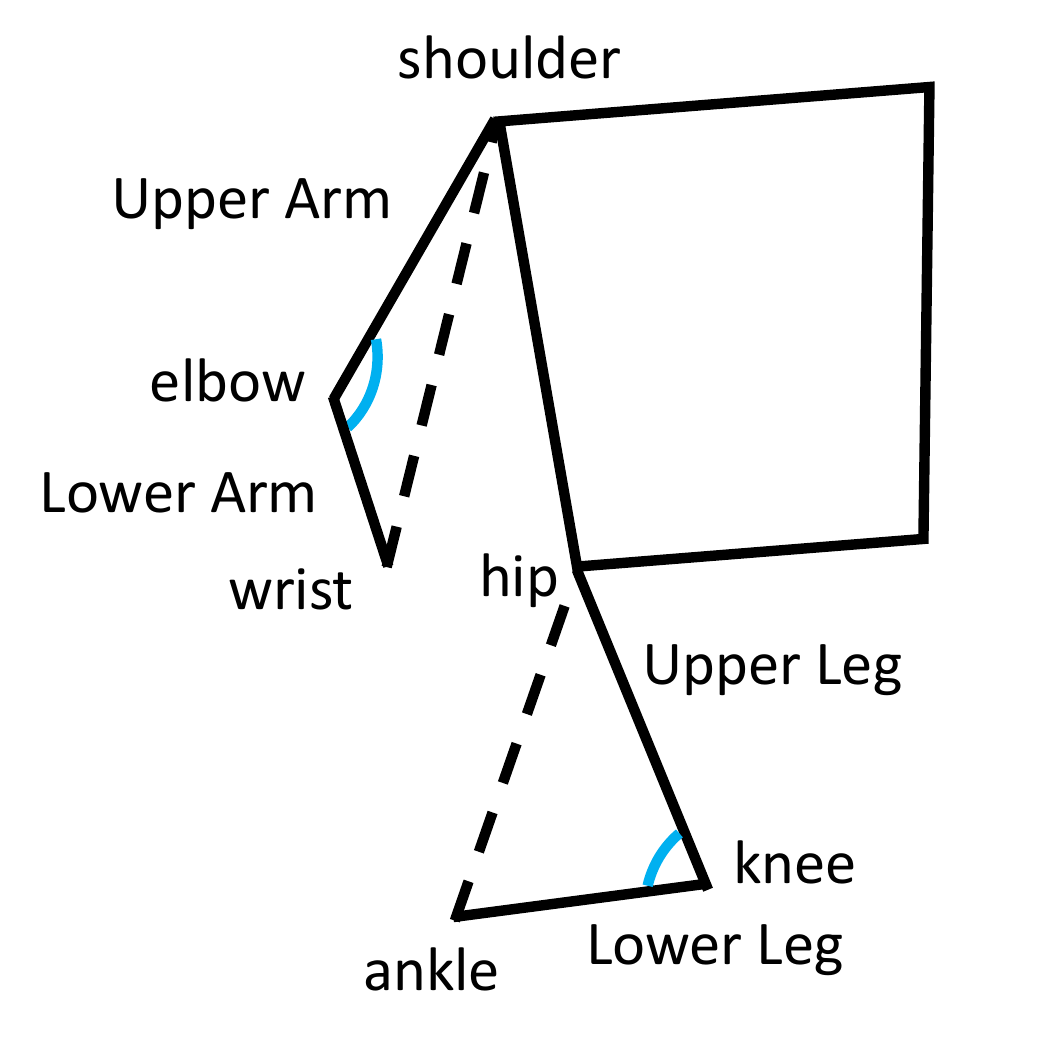}
    \caption{Arm/Leg Bent\label{fig:bent_feature}}
  \end{subfigure}
  \begin{subfigure}[b]{0.3\textwidth}
    \includegraphics[width=\textwidth]{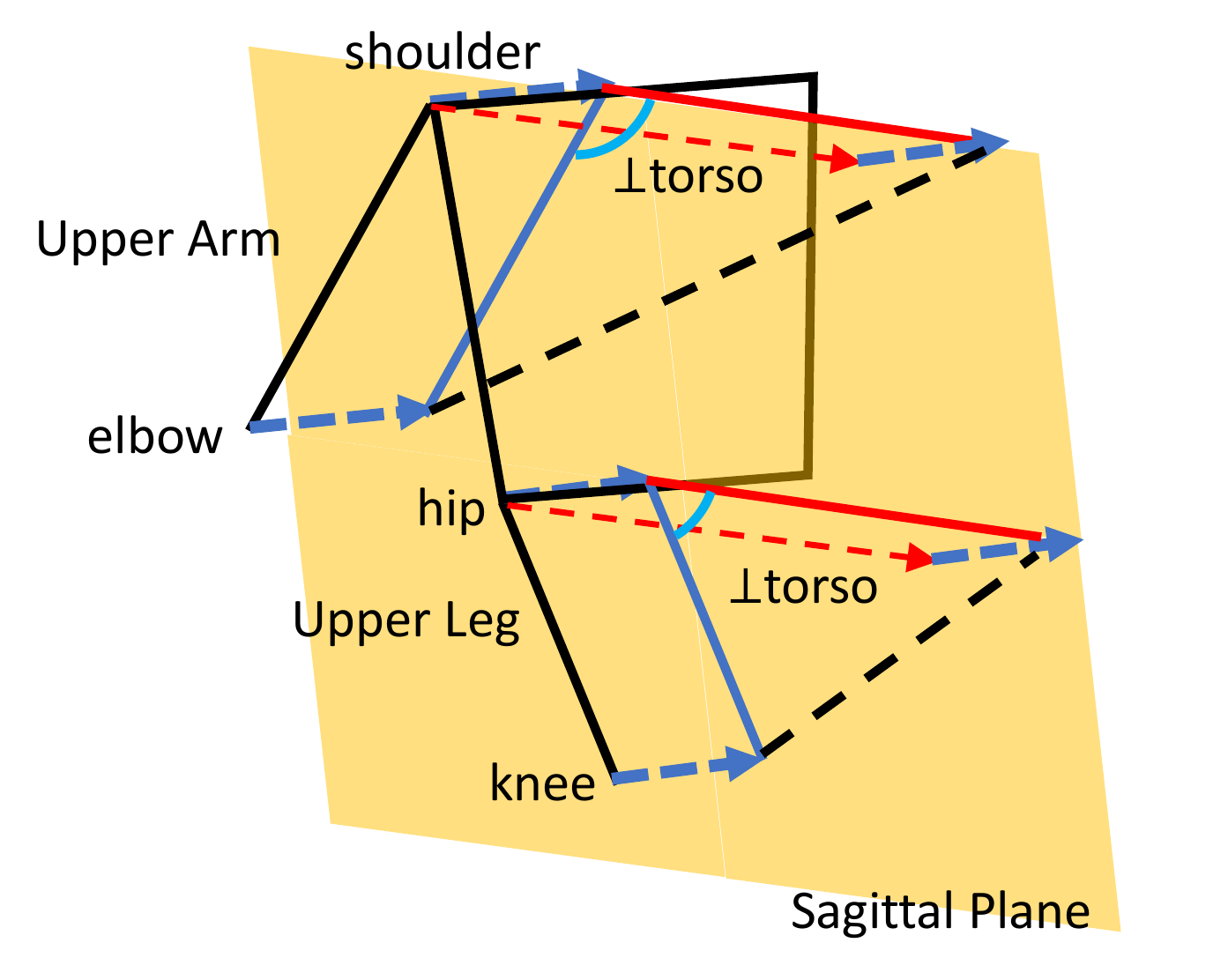}
    \caption{Arm/Leg In Front\label{fig:infront_feature}}
  \end{subfigure}
  \begin{subfigure}[b]{0.22\textwidth}
    \includegraphics[width=\textwidth]{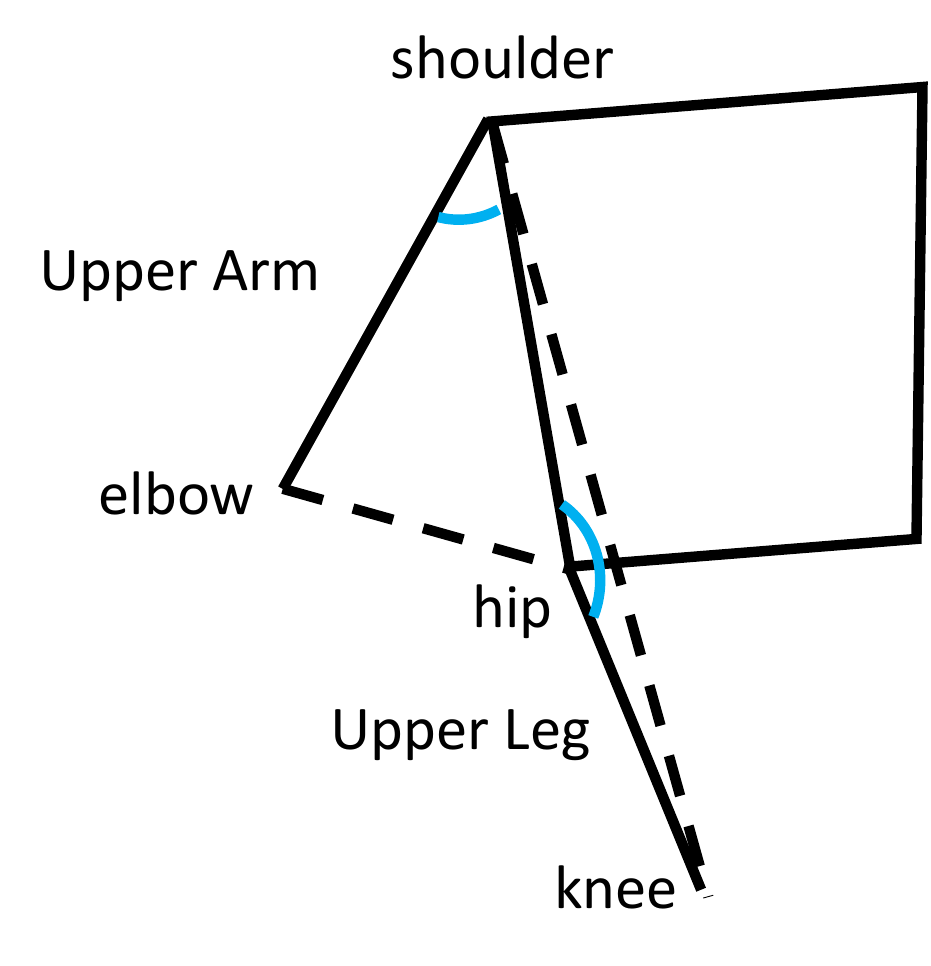}
    \caption{Arm/Leg Raised\label{fig:raised_feature}}
  \end{subfigure}
  \newline
  \begin{subfigure}[b]{0.28\textwidth}
    \includegraphics[width=\textwidth]{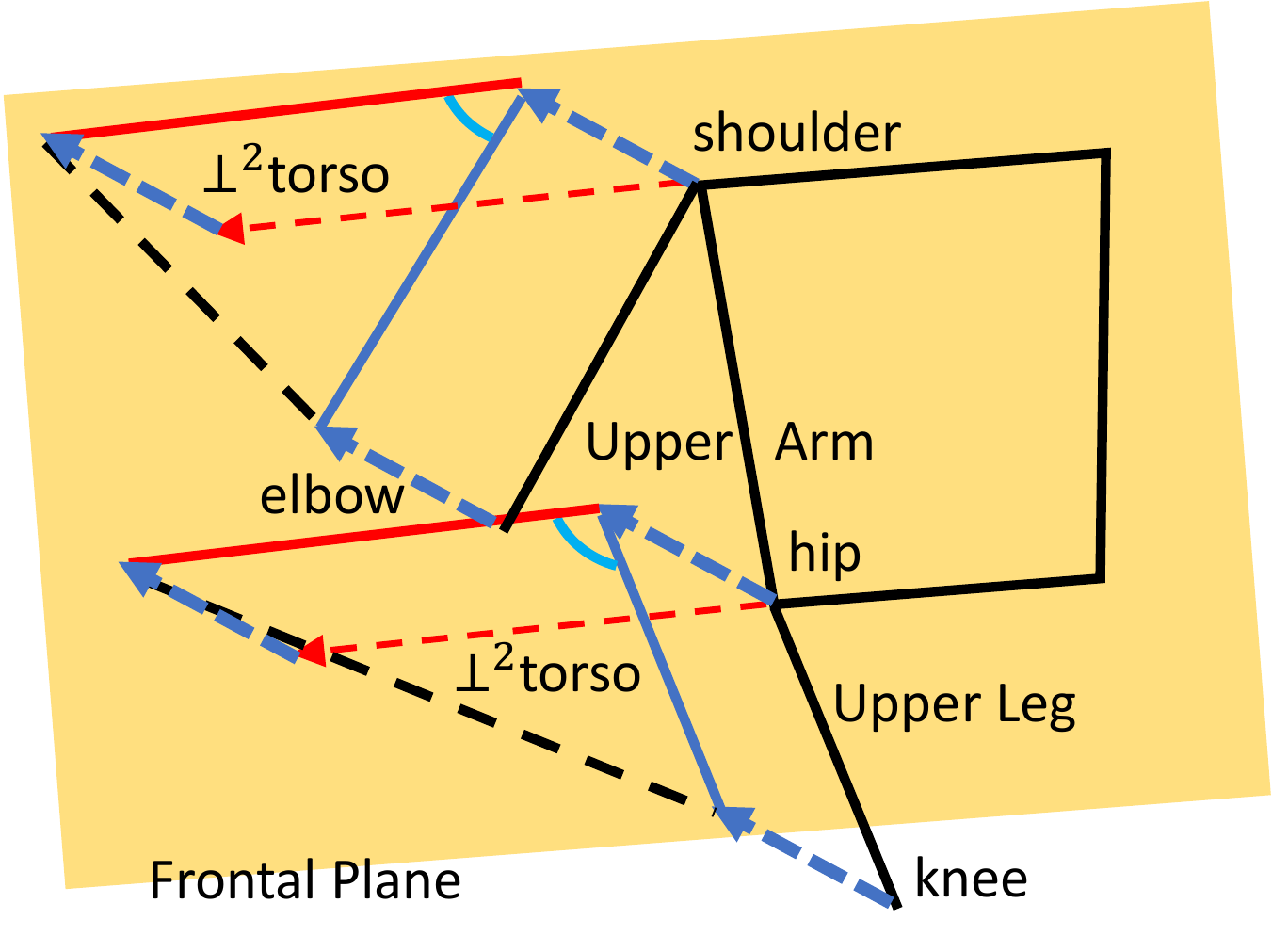}
    \caption{Arm/Leg Outward\label{fig:outward_feature}}
  \end{subfigure}
  \begin{subfigure}[b]{0.22\textwidth}
    \includegraphics[width=\textwidth]{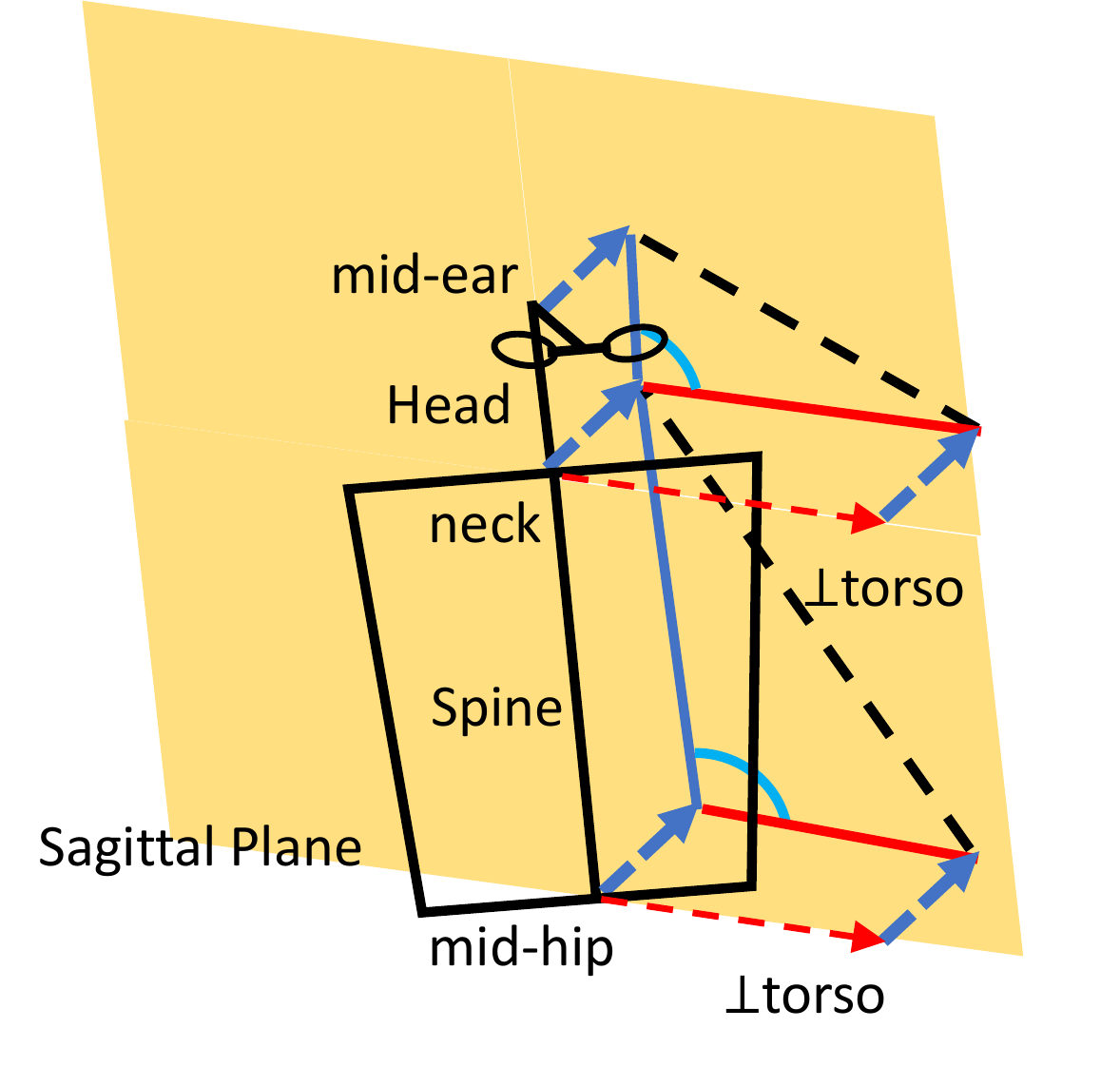}
    \caption{Head/Torso Tilted (Forward)\label{fig:tilted_forward_feature}}
  \end{subfigure}
  \begin{subfigure}[b]{0.22\textwidth}
    \includegraphics[width=\textwidth]{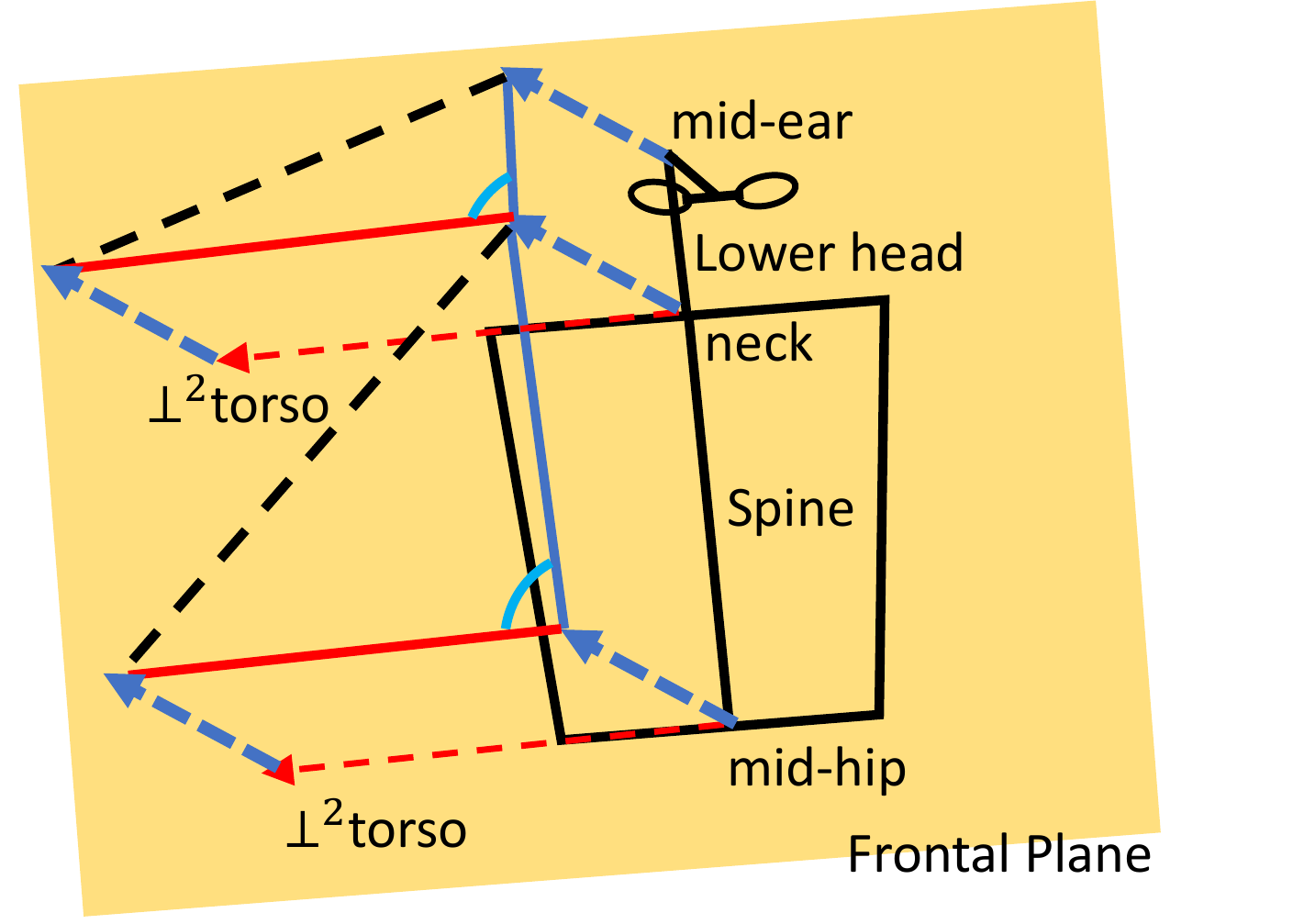}
    \caption{Head/Torso Tilted (Left)\label{fig:tilted_left_feature}}
  \end{subfigure}
  \begin{subfigure}[b]{0.25\textwidth}
    \includegraphics[width=\textwidth]{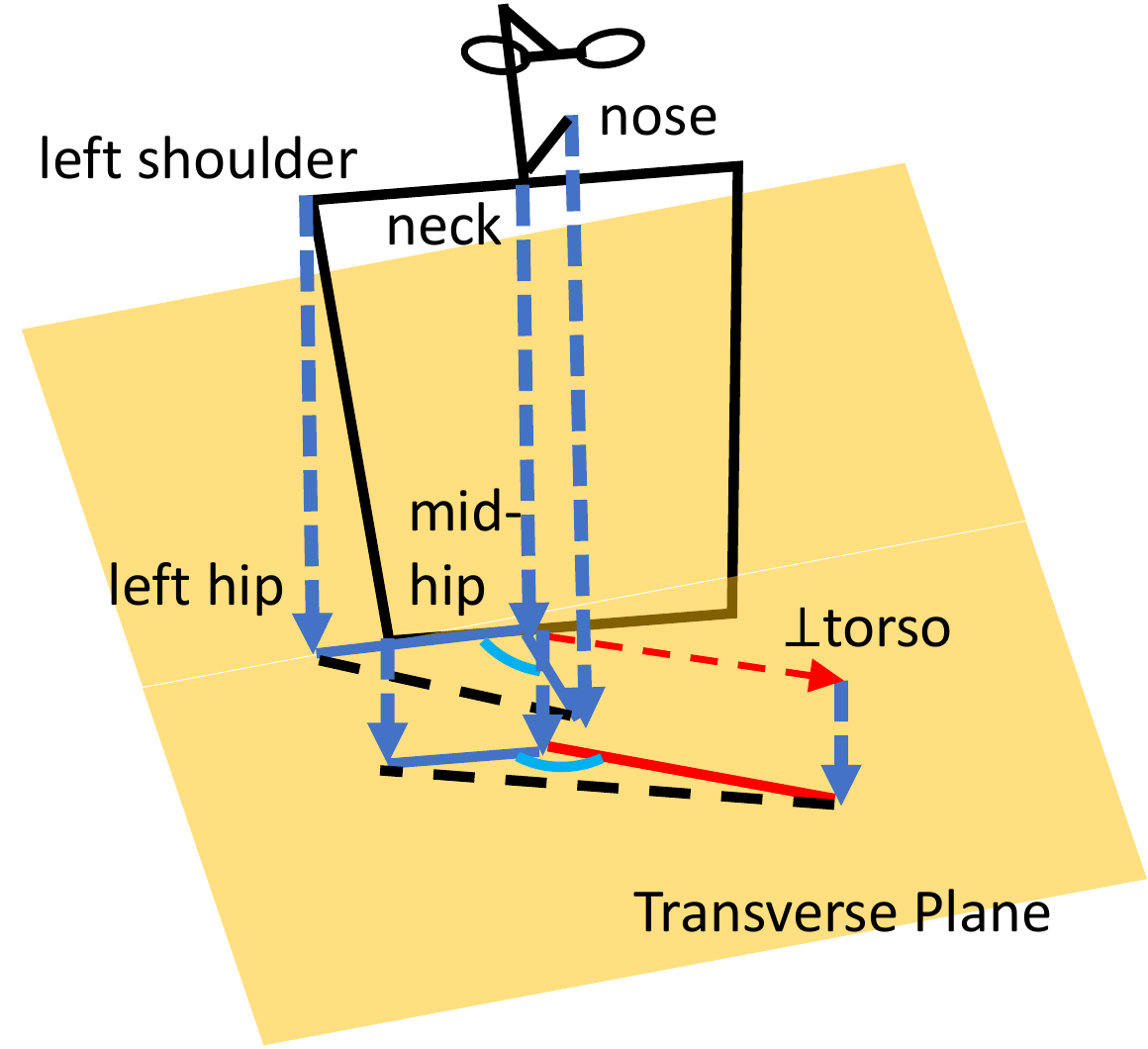}
    \caption{Head/Torso Twisted\label{fig:twisted_feature}}
  \end{subfigure}
  \caption{Illustrations of the triangles whose angles (teal) generate the fluent symbols.\label{fig:illustrated_features}}
\end{figure*}

\paragraph{Bent} describes the arms and legs with respect to the angles of the elbow and knee, respectively (see Figure~\ref{fig:bent_feature}).  The $\cdot$ side's elbow's angle applies Equation~\ref{eq:lawOfCos} to the triangle formed from links $\overrightarrow{\cdot\textnormal{ shoulder},\cdot\textnormal{ elbow}}$ and $\overrightarrow{\cdot\textnormal{ elbow},\cdot\textnormal{ wrist}}$.  The $\cdot$ side's knee's angle applies Equation~\ref{eq:lawOfCos} to the triangle formed from links $\overrightarrow{\cdot\textnormal{ hip},\cdot\textnormal{ knee}}$ and $\overrightarrow{\cdot\textnormal{ knee},\cdot\textnormal{ ankle}}$.  There are two states for the instantaneous fluent, $s_{0} = $ bent and $s_{1} = $ straight; the locally temporal counterparts are $\rightarrow s_{0} = $ bending and $\rightarrow s_{1} = $ straightening.  

\paragraph{In Front} describes the arms and legs with respect to the angles of the upper arm and leg when projected onto the sagittal plane \cite{knee-joint-applications}, respectively.  The reference line to which the upper appendages are compared is the torso plane's normal through the shoulder and hip, each facing in the forwards direction (see Figure~\ref{fig:infront_feature}).  The $\cdot$ side's upper arm's angle applies Equation~\ref{eq:lawOfCos} to the triangle formed from links $\textnormal{proj}_{\textnormal{sagittal}}\overrightarrow{\cdot\textnormal{ shoulder},\cdot\textnormal{ elbow}}$ and $\textnormal{proj}_{\textnormal{sagittal}}\overrightarrow{\cdot\textnormal{ shoulder},\cdot\textnormal{ shoulder}+\perp\textnormal{torso}}$ where $\perp\textnormal{torso} = \overrightarrow{\textnormal{left shoulder},\textnormal{left hip}}\times\overrightarrow{\textnormal{left shoulder},\textnormal{right shoulder}}$.  The $\cdot$ side's lower leg's angle applies Equation~\ref{eq:lawOfCos} to the triangle formed from links $\textnormal{proj}_{\textnormal{sagittal}}\overrightarrow{\cdot\textnormal{ hip},\cdot\textnormal{ knee}}$ and $\textnormal{proj}_{\textnormal{sagittal}}\overrightarrow{\cdot\textnormal{ hip},\cdot\textnormal{ hip}+\perp\textnormal{torso}}$.  There are three states for the instantaneous fluent, $s_{0} = $ in front, $s_{1} = $ centered, and $s_{2} = $ behind; the locally temporal counterparts are $\rightarrow s_{0} = $ moving forward and $\rightarrow s_{2} = $ moving backward.  

\paragraph{Raised} describes the arms and legs with respect to the angles of the armpit and hip from the torso without any projections (see Figure~\ref{fig:raised_feature}).  The $\cdot$ side's armpits's angle applies Equation~\ref{eq:lawOfCos} to the triangle formed from links $\overrightarrow{\cdot\textnormal{ shoulder},\cdot\textnormal{ elbow}}$ and $\overrightarrow{\cdot\textnormal{ shoulder},\cdot\textnormal{ hip}}$.  The $\cdot$ side's hip's angle applies Equation~\ref{eq:lawOfCos} to the triangle formed from links $\overrightarrow{\cdot\textnormal{ hip},\cdot\textnormal{ knee}}$ and $\overrightarrow{\cdot\textnormal{ hip},\cdot\textnormal{ shoulder}}$.  There are three states for the arm's instantaneous fluent, $s_{0} = $ lowered, $s_{1} = $ chest-level, and $s_{2} = $ raised, but the leg's instantaneous fluent can only be either $s_{0} = $ raised or $s_{1} = $ lowered.  The locally temporal counterparts for both the arms and legs are $\rightarrow s_{0} = $ raising and $\rightarrow s_{2} = $ lowering (for arms, $\rightarrow s_{1}$ for legs).

\paragraph{Outward} describes the arms and legs with respect to the angles of the upper arm and upper leg projected onto the frontal plane, 
respectively.  The reference line to which the upper appendages are compared is the normal of the sagittal plane, which we will call the frontal as it lies on the frontal plane, facing outwards (see Figure~\ref{fig:outward_feature}).  The $\cdot$ side's upper arm's angle applies Equation~\ref{eq:lawOfCos} to the triangle formed from links $\textnormal{proj}_{\textnormal{frontal}}\overrightarrow{\cdot\textnormal{ shoulder},\cdot\textnormal{ elbow}}$ and $\textnormal{proj}_{\textnormal{frontal}}\overrightarrow{\cdot\textnormal{ shoulder},\cdot\textnormal{ shoulder}+\perp^{2}\textnormal{torso}}$ where $\perp^{2}\textnormal{torso} = \overrightarrow{\cdot\textnormal{ shoulder},\cdot\textnormal{ hip}}\times\overrightarrow{\cdot\textnormal{ shoulder},\cdot\textnormal{ shoulder}+\perp\textnormal{torso}}$.  The $\cdot$ side's lower leg's angle applies Equation~\ref{eq:lawOfCos} to the triangle formed from links $\textnormal{proj}_{\textnormal{frontal}}\overrightarrow{\cdot\textnormal{ hip},\cdot\textnormal{ knee}}$ and $\textnormal{proj}_{\textnormal{frontal}}\overrightarrow{\cdot\textnormal{ hip},\cdot\textnormal{ hip}+\perp^{2}\textnormal{torso}}$.  
There are three states for the instantaneous fluent, $s_{0} = $ outward, $s_{1} = $ at side, and $s_{2} = $ inward; the locally temporal counterparts are $\rightarrow s_{0} = $ moving outward and $\rightarrow s_{2} = $ moving inward.

\paragraph{Tilted} describes the neck and mid-hip with respect to their angles in the sagittal or frontal plane.  The reference line to which the link is compared is either the torso plane's normal projected onto the sagittal plane or the torso plane's frontal projected onto the frontal plane (see Figures~\ref{fig:tilted_forward_feature} and~\ref{fig:tilted_left_feature}).  The neck's angle applies Equation~\ref{eq:lawOfCos} to two triangles.  One triangle for the forwards-direction contains links $\textnormal{proj}_{\textnormal{sagittal}}\overrightarrow{\textnormal{neck},\textnormal{mid-ear}}$ and $\textnormal{proj}_{\textnormal{sagittal}}\overrightarrow{\textnormal{neck},\textnormal{neck}+\perp\textnormal{torso}}$ where mid-ear $ = 0.5\left(\textnormal{left ear}+\textnormal{right ear}\right)$.  The other triangle for the leftwards-direction contains links $\textnormal{proj}_{\textnormal{frontal}}\overrightarrow{\textnormal{neck},\textnormal{mid-ear}}$ and $\textnormal{proj}_{\textnormal{frontal}}\overrightarrow{\textnormal{neck},\textnormal{neck}+\perp^{2}\textnormal{torso}}$.  The two triangles for the mid-hip's angles are similar, but substitute the following joints: neck becomes mid-hip where mid-hip  $ = 0.5\left(\textnormal{left hip}+\textnormal{right hip}\right)$ and mid-ear becomes neck.  For each direction, there are three states for the instantaneous fluent, $s_{0} = $ tilted forward, $s_{1} = $ centered, and $s_{2} = $ tilted backward; $s_{0} = $ tilted left, $s_{1}$ = centered, and $s_{2}$ = tilted right.  Due to how these fluents pair, the confidence of $s_{1}$ extends Equation~\ref{eq:confInstantaneous_sumTo1} using both instances of $s_{0}$ and $s_{2}$ per direction.  The locally temporal counterparts in each direction are $\rightarrow s_{0} = $ tilting forward or tilting left and $\rightarrow s_{2} = $ tilting backward or tilting right.

\paragraph{Twisted} describes the neck and mid-hip with respect to the angles in the transverse plane. 
The reference lines to which they are compared are the frontals projected onto the transverse plane (see Figure~\ref{fig:twisted_feature}).  The neck's angle applies Equation~\ref{eq:lawOfCos} to the triangle formed from links $\textnormal{proj}_{\textnormal{transverse}}\overrightarrow{\textnormal{neck},\textnormal{nose}}$ and $\textnormal{proj}_{\textnormal{transverse}}\overrightarrow{\textnormal{neck},\textnormal{left shoulder}}$.  The mid-hip's angle applies Equation~\ref{eq:lawOfCos} to the triangle formed from links $\textnormal{proj}_{\textnormal{transverse}}\overrightarrow{\textnormal{left hip},\textnormal{mid-hip}}$ and $\textnormal{proj}_{\textnormal{transverse}}\overrightarrow{\textnormal{mid-hip},\textnormal{mid-hip}+\perp\textnormal{torso}}$.  There are three states for the instantaneous fluent, $s_{0} = $ twisted left, $s_{1} = $ centered, and $s_{2} = $ twisted right; the locally temporal counterparts are $\rightarrow s_{0} = $ twisting left and $\rightarrow s_{2} = $ twisting right.

\paragraph{Near} describes how close any two joints in the stick figure representation are.  There are three states for the instantaneous fluent, $s_{0} = $ near, $s_{1} = $ far, and $s_{2} = $ touching.  Because distances are usually relative to one's stature, we use head-lengths as a unit of measurement; head-lengths are relatively consistent in human anatomy if patients have traditional proportions \cite{human-proportions-drawing}.  For our stick figure representation, we compute this unit measurement as $1$ headlength $ = 2\left|\overrightarrow{\textnormal{neck},\textnormal{mid-ear}}\right|$.  
To avoid an explosion of fluents describing the posture because there are many joints to consider, we restricted the possible pairings to $J_{from} = \lbrace$ left index finger, right index finger, left heel, right heel $\rbrace$ and $J_{to} = J_{from} \cup \lbrace$ nose, left eye, right eye, left ear, right ear, left mouth, right mouth, left shoulder, left elbow, left wrist, right shoulder, right elbow, right wrist, left hip, right hip, left knee, right knee, left big toe, right big toe $\rbrace$.  Due to tautologies when the some joints share links (their distances are relatively fixed), we omit some pairings from $J_{from} \times J_{to}$: ($\cdot$ index finger, $\cdot$ wrist), ($\cdot$ index finger, $\cdot$ elbow), ($\cdot$ heel, $\cdot$ knee), ($\cdot$ heel, $\cdot$ big toe), and the reflexive ($\cdot$, $\cdot$).  For all the permitted pairings, we simply defined the thresholds for near as $0.25$ headlengths or less and far as $0.75$ headlengths or greater with confidence $1$, which gives us $\tau^{I,F\left(\textnormal{distance}\left(j_{1} \in J_{from}, j_{2} \in J_{to}\right),\textnormal{near}\right)}_{0,1} = 0.5\textnormal{headlength}$ and $\delta^{I,F\left(\textnormal{distance}\left(j_{1} \in J_{from}, j_{2} \in J_{to}\right),\textnormal{near}\right)}_{0,1} = 0.25\textnormal{headlength}$.  We computed touching as $0.1$ distance or less \emph{without} accounting for headlength, and the certainty over this fluent is exclusively binary.  The locally temporal temporal counterparts are $\rightarrow s_{0} = $ approaching and $\rightarrow s_{1} = $ distancing with thresholds $\tau^{LT,F\left(\textnormal{distance}\left(j_{1} \in J_{from}, j_{2} \in J_{to}\right),\rightarrow\textnormal{near}\right)}_{0,1} = -0.05\textnormal{headlength}$, $\tau^{LT,F\left(\textnormal{distance}\left(j_{1} \in J_{from}, j_{2} \in J_{to}\right),\rightarrow\textnormal{far}\right)}_{0,1} = 0.05\textnormal{headlength}$, and $\delta^{LT,F\left(\textnormal{distance}\left(j_{1} \in J_{from}, j_{2} \in J_{to}\right),\rightarrow\textnormal{near}\right)}_{0,1} = \delta^{LT,F\left(\textnormal{distance}\left(j_{1} \in J_{from}, j_{2} \in J_{to}\right),\rightarrow\textnormal{far}\right)}_{1,0} = 0.05\textnormal{headlength}$.

\subsection{High Level: Qualitative Motion Primitives\label{sec:approach.qmp}}
Using the angles computed above, we also define more general patterns of motion to describe human posture over longer periods of time than a few frames.  Similar to the change in complexity from a regular expression compared to a grammar, we can express richer relationships as parts of the human body change their posture in structured ways.  Unlike traditional motion primitives that represent a set of possible robot movements in Cartesian space \cite{DBLP:conf/icra/CohenCL10}, we present qualitative motion primitives (QMPs) that describe 
patterns of human movement.

The characterization of QMPs uses Sparse Identification of Non-linear Dynamics (SINDy) \cite{brunton2016discovering,fasel2021sindy}.  SINDy has effectively reconstructed diverse types of dynamic models directly from time-series state data.  It uses a sequentially thresholded least-squares algorithm to identify a sparse subset of basis functions that best fit the observed state evolution.  The method is not computationally intensive, and it exhibits a sharp rate of convergence on a wide range of dynamic structures \cite{zhang2019convergence}.  These make it an attractive candidate for real-time applications of model learning.

In general, we can express a dynamic system analytically in the following form.  The states of our system are composed primarily of the angles between limbs and the angles formed between a limb and a body-fixed axis, which Figure~\ref{fig:illustrated_features} illustrates.  Each state derivative is a function of the states and controls:
\begin{equation}\label{eq:dynsys}
 \dot{\bm{X}} = \bm{\Phi}(\bm{X},\bm{U}) \bm{\Lambda} 
\end{equation}
where $\bm{X}$ is a vector of states, $\bm{U}$ is a vector of controls, $\bm{\Phi}(\bm{X},\bm{U})  = \left[  \phi_1(\bm{X},\bm{U}),  \ldots,  \phi_n(\bm{X},\bm{U}) \right]$ is a set of candidate  basis functions for  the right-hand-side dynamics, and $ \bm{\Lambda} = [\bm{\lambda}_1, \bm{\lambda}_2,\ldots,  \bm{\lambda}_n]$ is the set of real coefficients on those basis functions.

Given a time history of controls $\bm{U}$, estimated states $\bm{X}$, and estimated state derivatives $\dot{\bm{X}}$, the model-learning problem finds the matrix of coefficients $\bm{\Lambda}$ that both 1) minimizes the residual error between observed states and what the model predicts, and 2) promotes sparsity in $\bm{\Lambda}$.  The two conditions for $\bm{\Lambda}$ represent competing objectives: the desire for accuracy vs. the desire to express the model using a minimal number of terms.
For the $k^{\textnormal{th}}$ state, the problem becomes:
\begin{equation}\label{dynsys_kthstate}
\bm{\lambda}_k = {arg}\min_{\hat{\lambda}_k} || \dot{\bm{X}}_k - \bm{\hat{\lambda}}_k \Phi^T(\bm{X},\bm{U})||_2 + \alpha||\bm{\hat{\lambda}}_k||_1
\end{equation}
where we set parameter $\alpha$ to find a Pareto optimal solution that provides a desired balance between high accuracy and low model complexity.

We assume that we can model each state as an explicit function of time and other states to describe basic human motions.  This represents a special case where there is only one control signal, defined simply as the time vector $\bm{U}=t$.  Thus, in our formulation, the basis functions forming all QMPs are represented as $\Phi(\bm{X},t)$. 
We also assume that our hand-picked selection of candidates for the set of basis functions $\Phi$ is sufficient to solve the specific instance of the model-learning problem: 
\begin{eqnarray}
\phi_1(\bm{X},t) &=& 1 \label{eq:primitive-basis-1}\\
\phi_2(\bm{X},t) &=& t  \label{eq:primitive-basis-2}\\
\phi_3(\bm{X},t) &=& \cos(2\pi t/T)  \label{eq:primitive-basis-3}\\
\phi_4(\bm{X},t) &=& \sin(2\pi t/T)  \label{eq:primitive-basis-4}\\
\phi_5(\bm{X},t) &=& \cos(\pi t/T)   \label{eq:primitive-basis-5}\\
\phi_6(\bm{X},t) &=& \sin(\pi t/T) . \label{eq:primitive-basis-6}
\end{eqnarray}

QMP characterization considers a range of periods $T$ when solving the regression problem for $\Lambda$, and it chooses the $T$ that gives the smallest residual.  We intentionally choose sinusoidal basis functions with both the whole- and half-period to account for the combined presence of long- and short-period motion.

It is worth noting that each of the six basis functions is simply an explicit function of time, rather than a more general state-dependent ordinary differential equation (ODE).  The above functions have shown to be sufficient in characterizing several of the basic motions we examined thus far.  However, as future applications use this representation to evaluate more types of motion for their domain, it might become necessary to add additional basis functions, including state-dependent ODEs.  Expanding the set of basis functions ensures that the important characteristics of the limb and joint articulation are appropriately captured.  Because many activities involve coordinated and synchronized motion among multiple limbs and joints, we anticipate that first- or second-order ODEs may serve as useful models.

As parametric models that describe the skeletal state evolution over a given time window, we then evaluate them against the criteria associated with each entry in an Action Description Database (ADD).  This database encodes our QMPs based on medical expert-generated descriptions.  It is simply a repository of functions with thresholds, each representing a fluent that maps the results of the characterized QMPs to a binary value.  This evaluation results in a numeric score for each entry in the ADD, indicating how well the parametric model meets various criteria.  One important benefit of this approach is that we do not assume people exclusively perform only one QMP at a time.  Rather than choose the greatest-scoring one at the expense of others, this approach considers the possibility that people perform multiple motion primitives simultaneously.

\section{Interpretable Activity Recognition\label{sec:evaluation}}
\begin{figure*}
  \centering
  \includegraphics[width=\textwidth]{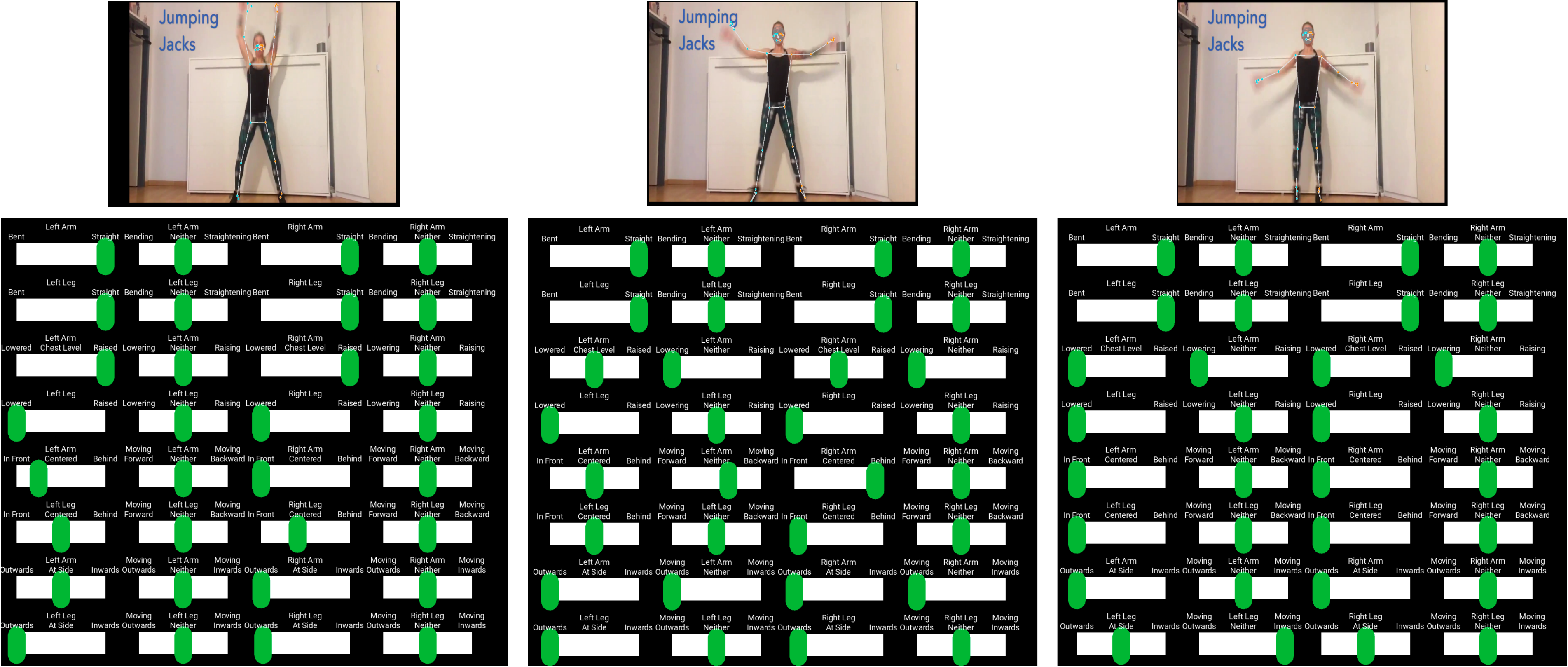}
  \caption{Fluent descriptions for frames of a jumping jack animation (video source: Kinetics 700 Dataset \cite{carreira2019short}).\label{fig:postureDescribed}}
\end{figure*}

Figure~\ref{fig:postureDescribed} illustrates how the representations in Section~\ref{sec:approach.descriptors} successfully describe the human posture in a way that many people would.  
A collection of postures over time with these descriptions contains trends that further describe how a person physically performs various activities.  We created two activity recognition classifiers, one encoded 
by hand and the other autonomously generated, 
as a preliminary empirical evaluation of our proposed representation's interpretability within AI systems.  See \citeauthor{blmjwwf_xcbr2022} \shortcite{blmjwwf_xcbr2022} for more details.

\subsection{Dynamic Action Detection\label{sec:evaluation.dad}}
Dynamic Action Detection (DAD) detects pre-defined activities from time-series data of skeletal motion.  
We focus specifically on the time-varying states of various body parts and attempt to match them to one or more QMPs.  DAD 
is versatile because it provides solutions based on pre-defined motion descriptions.  When the QMPs that make up an activity are known and well-defined, then activity recognition can be configured as an expert system without the need for any training data.  Alternatively, given a broad enough set of metrics \emph{and} properly labeled videos from a dataset, then the descriptive features throughout Section~\ref{sec:approach} may serve as inputs for a machine learning algorithm to identify which features are the best descriptors for different activities.


We developed a 
questionnaire 
to gather detailed descriptions on four different activities from our medical experts: squat, push-up, jumping jack, and a single leg Romanian deadlift.  For each activity, we asked experts to describe all the bodily motion attributes (our QMPs) that are important to successfully execute that action.  The questionnaire also asks experts to specify a weight between $1$ and $10$ to rank the relative importance of each attribute.  We used the results 
as guidance to develop our ADD.

\begin{figure}
  \centering
  \includegraphics[scale=0.9]{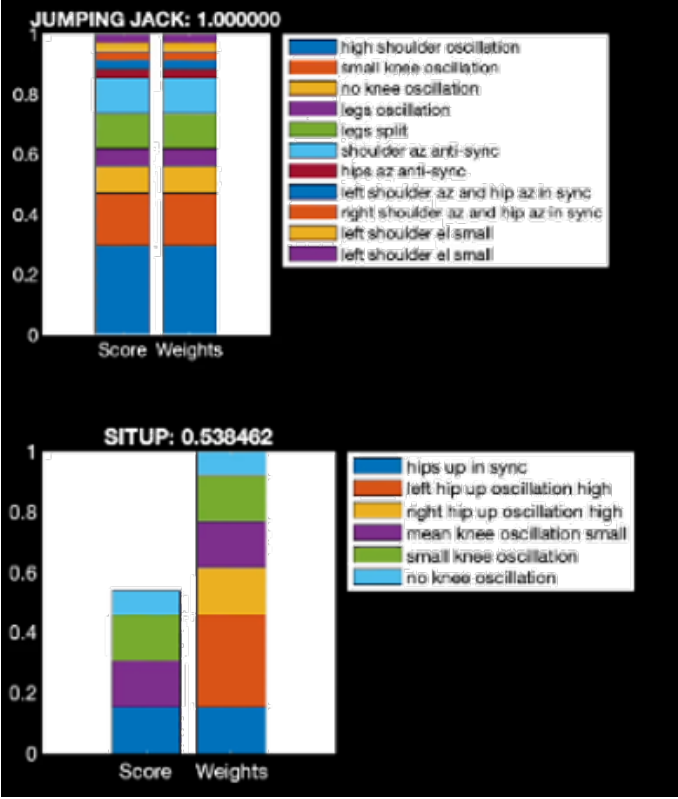}
  \caption{DAD compares the ratio of detected QMP patterns over a time window per activity.  
    The model is on the right.
    \label{fig:dadActivityBreakdown}}
\end{figure}

With this instance of the database, we 
created an expert system for activity recognition that defines activities hierarchically as
a sequence of two or more QMPs. 
As DAD runs and successively identifies multiple QMPs across time, it can also look for sequences of them that comprise these activities. 
For example, our current version of DAD can identify knee oscillation, synchronized shoulder translation, and consistent leg distance (see Figure~\ref{fig:dadActivityBreakdown}).
If these actions are detected in the correct sequence according to the proper criteria, then the sequence could be identified as a squat. 
DAD will list the QMPs that are executed as expected for each activity as well as those that are not.  This listing serves as an interpretable model for people to inspect, understand, and validate how DAD performs activity recognition.

\subsection{Case-Based Reasoning\label{sec:evaluation.cbr}}
Case-based reasoning \cite{kolodner1992introduction} (CBR) is a learn-by-experience algorithm that compares new inputs to previously-seen examples.  Our CBR approach for activity recognition stores examples of people performing activities from videos using one of the qualitative posture representations in Section~\ref{sec:approach}.  From a supervised machine learning perspective, the set of features describing previously-seen activity instances is called the problem; the classification labels for the activity are called the solution.  The learned model is a case-base (CB) data structure that maintains a set of cases, which pair previously-seen problems with their solutions.

The classification step compares the problem representing the currently-observed video, converted into the same qualitative representation, against cases within the CB to determine which of the previously-seen instances of each activity are most similar.  This differs from $k$-nearest neighbors \cite{cover1967nearest} because CBR adds additional learning steps for revision and retention \cite{smyth1995remembering}.  Revision happens during classification, adjusting the most similar case's problem and solution to match the current world state.  Retention happens after classification, adding the newest observed problem and its recognized solution to the CB.  To manage the size of the CB as it grows, 
we consider the novelty and benefit of adding the case to the CB \cite{borck2017automated}.  This will trim the CB when initialized and regulate which new cases are added, reducing 
bias that stems from too many similar cases.

\begin{figure}
\centering
    \includegraphics[angle=-90,scale=0.33]{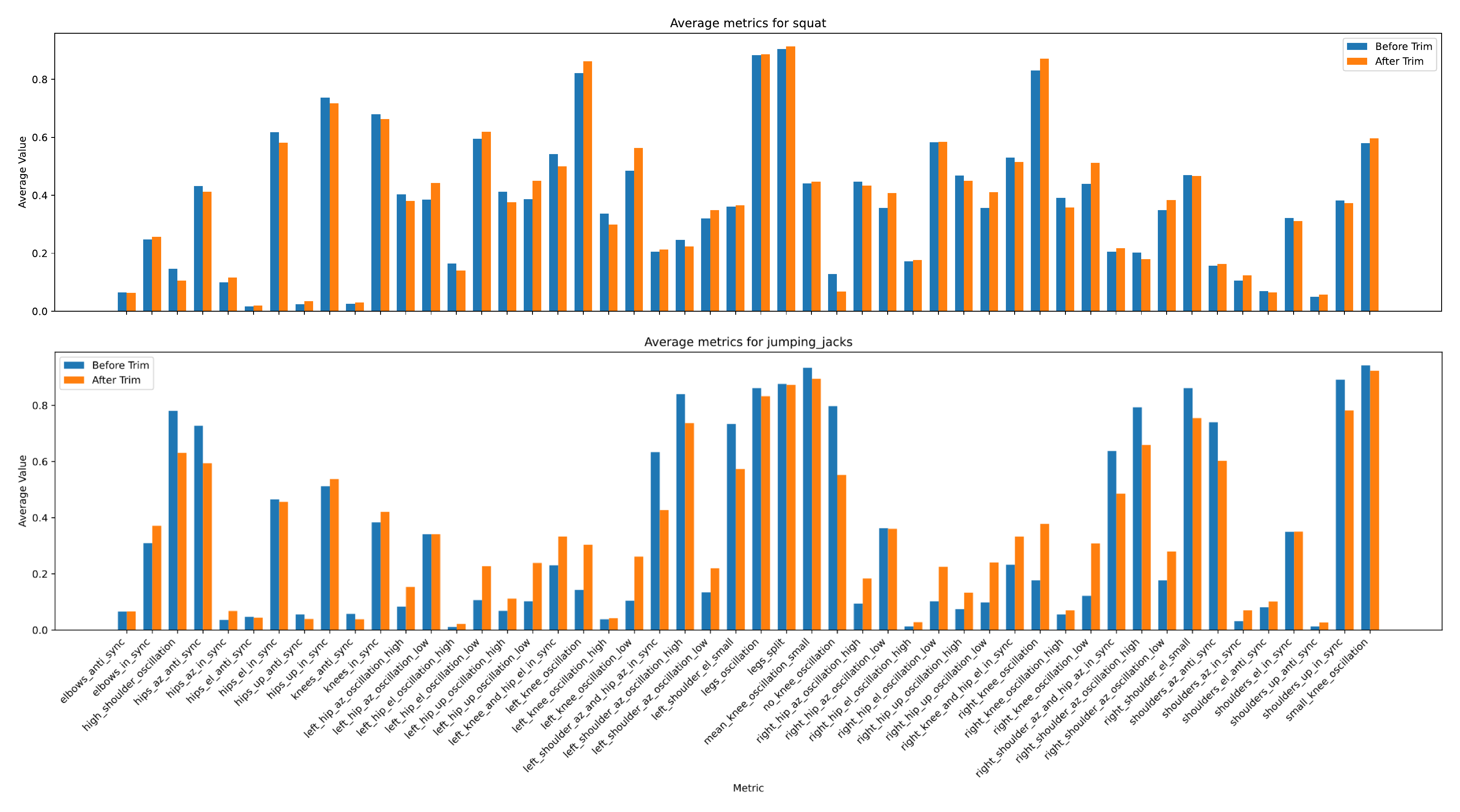}
    \caption{Learned features over squat vs. jumping jack cases in the initial and trimmed CB.\label{fig:validation-average-metrics}}
\end{figure}

We tried several different similarity functions for problem comparison: Euclidean distance, Dynamic Time Warping \cite{sakoe1971dynamic}, and signatures \cite{lyons1998differential}.  Our experiments presented poor accuracy for all these metrics when using the entire set of fluent descriptions (Section~\ref{sec:approach.descriptors}) at every timestep.  We hypothesize that considering this level of descriptive detail is too fine-grained to learn effective patterns \cite{bellman1961adaptive}.  Experiments using the QMPs (Section~\ref{sec:approach.qmp}) provide evidence supporting this hypothesis as the accuracy drastically improved with a simple Euclidean distance similarity measure, which counts the number of matched features in the case.
The features from our representation enable the case and comparison metrics to be interpretable, and we see distinct differences in the prevalence of features between different activities.  The CB trimming process retains a comparable distribution of features for each activity; see Figure~\ref{fig:validation-average-metrics}. 

\subsection{Generating Explanations\label{sec:evaluation.extensions}}
Both DAD and CBR perform activity recognition through a comparison of features that qualitatively describe the human posture in ways that are human-readable and quickly understandable.  This allows the model behind the algorithm to be transparent if people want to inspect how the robot performs activity recognition.  However, interpretability does not promise perfect performance---people will still experience the robot make mistakes and fail to understand what justifies some of its activity classifications.  These situations require explanations that clearly inform how the robot performed activity recognition for that specific instance.  The interpretable model invites a simple explanation that walks through the algorithm, stating which descriptions of the person's posture were observed and whether or not they match thoose associated with each activity in the model.

Outside of queries answering why the robot recognized a specific activity, our qualitative descriptions' roots in human semantics present the opportunity for robots to also explain how a person performs the activity.  Depending on the robot's application(s), this can serve as feedback to people or a form of introspection on the robot's own model.  For example, the activity recognition's relative scoring method might indicate that someone performing a recognized action is doing so with a lower score than desired for classification.  This indicates that either the observed person is not performing some attribute of that action very well or the model is missing some alternative way people perform the activity.  Explanations that discuss the scores associated with each feature enable people, be it the one observed or someone granted access to their data (caretaker, guardian, etc.), to examine what is wrong at a glance.

\section{Discussion and Future Work\label{sec:conclusions}}
There are infinitely many ways to represent streams of sensor data, but only so many of these are practical for robots and humans, balancing AI computational needs with what makes sense to the people who interact with these machines.  In contrast to many popular representations that focus towards the former, we present two related representations that bring the balance back towards the latter.  Starting with a stick figure representation of human posture, we encode angles between links into symbols that match qualitative descriptions that humans would typically use.  At the lower level, these are simple fluents that are myopic within a few frames; higher levels of abstraction identify QMPs that organize these relations into basic patterns of motion.  We presented evidence supporting these representations' interpretability with respect to the observed data and two activity recognition approaches, one hand-generated and the other machine-generated.  As AI-HRI involves people at a more direct level, we encourage other researchers to consider their representation choices and identify symbolic abstractions to extract before sending the data to AI algorithms, ensuring the information already has human-interpretable semantics.

There are still many things to study, both for other data representations and the ones we proposed.  In this paper, we manually set all the parameter values mapping angles to their qualitative relations.  This is not scalable as we define more symbols, and is one reason learning models became more sustainable laborwise compared to expert-derived ones.  More importantly, there is a risk that our choices are not representative of how others perceive the qualitative relations, biasing the symbol interpretations.  In future work, we plan to study at what values various people perceive them and determine if we can learn the parameters. 
We will also investigate connections between fluent descriptions and QMPs because both derivations stem from the same angle data.  A hierarchy of representations could refine interpretable models and explanations with respect to various levels of granularity.
Lastly, we will continue to identify additional fluent descriptions and QMPs for our proposed representations.  There are other parts of the body and ways that they move, adding more expressivity. 
Some of the derivations likely generalize to describe similar joint or link structures, which could extend the representation to other anatomies, including the robot's, for a more universal description that both humans and robots can interpret.

\section{Acknowledgments}
The authors thank the anonymous reviewers for their feedback to improve the quality of this paper.  This project has been funded in whole or in part with Federal funds from the National Cancer Institute, National Institutes of Health, Department of Health and Human Services, under Contract No.~75N91021C00039.

\bibliography{chacha-aihri}

\end{document}